%% file: main.tex
\documentclass[10pt,twocolumn,letterpaper]{article}

\usepackage[pagenumbers]{cvpr}   

\input{preamble}

\definecolor{cvprblue}{rgb}{0.21,0.49,0.74}
\usepackage[breaklinks,colorlinks,allcolors=cvprblue,
  pdftitle={Spoofing the Latent State: Imperceptible Goal-Image Attacks Derail Planning in LeWorldModel},
  pdfauthor={Kunal Bhosikar, Nicole Meng, Charu Sharma, Yingjie Lao}]{hyperref}

\def\confName{CVPR}
\def\confYear{2027}

\title{Spoofing the Latent State: Imperceptible \\ Goal-Image Attacks Derail Planning in LeWorldModel}

\author{%
Kunal Bhosikar$^{1, 3}$\hspace{1.3em}Nicole Meng$^{2}$\hspace{1.3em}Charu Sharma$^{1}$\hspace{1.3em}Yingjie Lao$^{2}$\\[0.45em]
{\normalsize $^{1}$IIIT Hyderabad\hspace{1.6em}$^{2}$Tufts University\hspace{1.6em}$^{3}$Miko}\\[0.2em]
{\footnotesize\tt kunal.bhosikar@research.iiit.ac.in, charu.sharma@iiit.ac.in, \{ziyi.meng, yingjie.lao\}@tufts.edu}
}

\begin{document}

\twocolumn[{%
\renewcommand\twocolumn[1][]{#1}%
\maketitle
\vspace{-1.6em}
\begin{center}
\centering
\includegraphics[width=\textwidth]{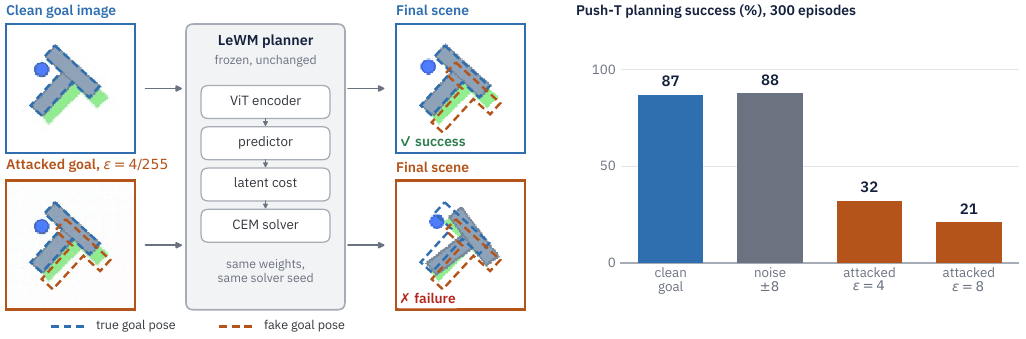}
\captionof{figure}{\textbf{A $4/255$ perturbation of the goal image redirects an unchanged
world-model planner.} The same Push-T episode is planned twice with identical LeWM weights, solver
and seed: with the clean goal image (top) and with one perturbed by at most $4/255$ per pixel
(bottom), visually indistinguishable from it. Dashed outlines mark the true goal pose (blue) and
the fake pose the attacker steers the encoder toward (orange): the clean run ends on the former,
the attacked run next to the latter (an episode in which the attack succeeds, \cref{app:examples};
scenes cropped to the region where the block moves). Right: planning success over $300$ episodes;
noise of twice the magnitude is harmless.}
\label{fig:teaser}
\end{center}%
\vspace{0.4em}
}]

\input{sec/0_abstract}
\input{sec/1_intro}
\input{sec/2_related}
\input{sec/3_background}
\input{sec/4_geometry}
\input{sec/5_method}
\input{sec/6_experiments}
\input{sec/7_conclusion}
{
    \small
    \bibliographystyle{ieeenat_fullname}
    \bibliography{main}
}

\clearpage
\appendix
\renewcommand{\thetable}{A\arabic{table}}\renewcommand{\theHtable}{A\arabic{table}}
\renewcommand{\thefigure}{A\arabic{figure}}\renewcommand{\theHfigure}{A\arabic{figure}}
\setcounter{table}{0}\setcounter{figure}{0}
\input{sec/8_appendix}

\end{document}

%% file: preamble.tex
\usepackage{pifont}
\usepackage{colortbl}
\newcommand{\cmark}{\textcolor{green!55!black}{\ding{51}}}
\newcommand{\xmark}{\textcolor{red!70!black}{\ding{55}}}
\DeclareMathOperator*{\argmin}{arg\,min}

\usepackage{microtype}

\makeatletter
\renewcommand\paragraph{\@startsection{paragraph}{4}{\z@}{1.1ex \@plus .4ex \@minus .2ex}{-1em}{\normalfont\normalsize\bfseries}}
\makeatother



%% file: sec/0_abstract.tex
\begin{abstract}
Joint-embedding predictive architectures (JEPAs) have recently emerged as general-purpose visual world models for control. 
A prominent example is LeWorldModel (LeWM), which trains end-to-end from pixels using an isotropic Gaussian anti-collapse regularizer (SIGReg) and plans by minimizing the squared Euclidean distance between predicted and goal embeddings. 
In this work, we investigate the test-time robustness of this latent planning interface. 
Analyzing the learned metric on Push-T, we first find that task-relevant object state heavily dominates latent distance: shifting the block by a single pixel displaces the embedding $3\times$ further than an equivalent agent shift, and swapping the block pose alone accounts for $99\%$ of the distance between scene pairs. 
However, because this pose information is distributed across a curved, high-dimensional manifold rather than a fixed linear subspace, the encoder cannot be trivially edited post-hoc to isolate or redirect it. 
We next demonstrate that this geometry is highly vulnerable at test time.
A white-box $\ell_\infty$ perturbation of just $\epsilon{=}4/255$ applied to a \emph{single} input frame causes the encoder to register the block at an arbitrary, attacker-chosen pose in $73$--$95\%$ of scenes without disturbing the encoded agent state. 
Perturbing either the goal image or the live camera observation drops Push-T planning success from $87\%$ to $32\%$, whereas uniform random noise at twice the magnitude has negligible impact. 
This vulnerability persists across all four standard LeWM benchmarks (Push-T, Two-Room, Reacher, and OGBench-Cube).
At $\epsilon{=}8/255$, the attack bridges $82$--$97\%$ of the squared latent distance to the target embedding in every environment, inducing planning degradation proportional to each task's reliance on the spoofed entity.
We also show that evaluating these attacks faithfully requires on-manifold counterfactual targets, as naive off-manifold re-renders can both fabricate and mask failure modes.
Because the attack operates purely in pixel space without modifying model weights, standard model-side diagnostics including SIGReg statistics, forward prediction error, and latent ``surprise'' monitors fail to detect it; goal-image perturbations bypass surprise checks entirely since the planner never predicts the goal state. 
Together, our findings expose the pixel-to-latent mapping in Gaussian-regularized JEPA planners as a critical attack surface and demonstrate that anti-collapse regularization during training confers no test-time adversarial robustness.
\end{abstract}

\begin{figure*}[t]
\centering
\includegraphics[width=\linewidth]{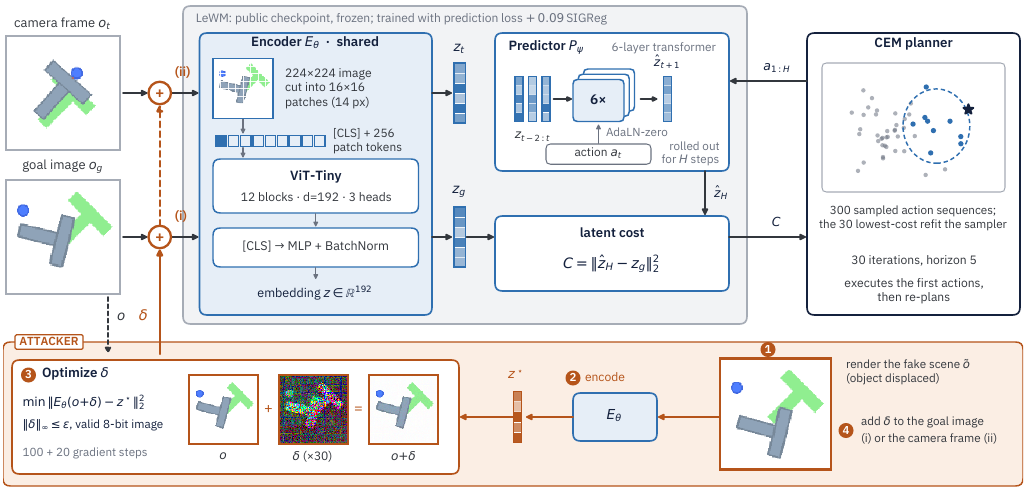}
\caption{\textbf{Overview.} Top: LeWM encodes the camera frame $o_t$ and the goal image $o_g$ with
a frozen ViT-Tiny encoder $E_\theta$, forecasts the latent $\hat z_H$ under candidate actions with
an action-conditioned transformer predictor $P_\psi$, and a CEM planner searches for the actions
that minimize the latent cost $C$. Bottom: our attacker (orange) leaves all of this untouched. It
(1) renders a counterfactual scene $\tilde o$ in which the task object is displaced, (2) encodes it
with the public encoder to a target $z^\star$, (3) optimizes an $L_\infty$-bounded perturbation
$\delta$ with $E_\theta(o{+}\delta)\approx z^\star$ by gradient steps through the encoder, and (4)
adds $\delta$ to the goal image~(i) or to the camera frames~(ii). The planner then pursues
the fake goal.}
\label{fig:overview}
\end{figure*}

%% file: sec/1_intro.tex
\section{Introduction}
\label{sec:intro}

World models allow agents to plan by predicting how an environment evolves across candidate future trajectories~\cite{ha2018worldmodels,hafner2025dreamerv3}. 
An increasingly popular approach instantiates these models as Joint-Embedding Predictive Architectures (JEPAs)~\cite{lecun2022path,assran2023ijepa}, where an encoder maps observations into a compact latent space and a predictor forecasts subsequent latents conditioned on actions without pixel reconstruction.
Because pretrained JEPAs are expensive to train and are now shared as off-the-shelf dynamics backbones~\cite{zhou2024dinowm,bardes2024vjepa}, their test-time reliability is a practical security concern as much as a benchmark accuracy problem.

LeWorldModel (LeWM)~\cite{maes2026lewm} is the first JEPA trained stably end-to-end from raw pixels using only a next-embedding prediction loss and the Sketched-Isotropic-Gaussian Regularizer (SIGReg)~\cite{balestriero2025lejepa}, which prevents collapse by constraining the latent distribution toward an isotropic Gaussian. 
At test time, LeWM plans via model-predictive control (MPC) by encoding the current observation and a goal image, rolling the predictor forward under candidate action sequences, and selecting the sequence whose terminal predicted embedding is closest in squared Euclidean distance to the goal.
Every control decision thus depends entirely on two encoder passes and a single latent distance metric.

This design exposes a natural vulnerability.
Because the planner observes only latent embeddings rather than true physical state, an adversary who subtly perturbs the input pixels to alter the encoded representation can cause the planner to optimize toward an incorrect outcome while leaving the model weights, solver, and goal specification intact \cite{mengwhispersplat}.
While adversarial examples are well studied for classifiers and generative models ~\cite{szegedy2014intriguing,goodfellow2015explaining,madry2018towards, meng2025advancing} and policy networks~\cite{huang2017adversarial,lin2017tactics,zhang2020robust}, JEPA planners differ in \textbf{two key respects}.
First, the perturbation must act through an \emph{intervening optimization loop}, reshaping the planner's cost landscape rather than directly perturbing action logits.
Second, LeWM is equipped with an isotropic Gaussian regularizer intended to regularize latent geometry and a runtime ``surprise'' score designed to flag out-of-distribution events~\cite{maes2026lewm}, inviting the question of whether these components provide built-in adversarial robustness.

We show that neither component protects the planner at test time.
We first examine how task-relevant state is represented in LeWM's latent metric on Push-T (\cref{sec:geometry}).
By rendering counterfactual scenes that vary one state factor at a time, we find that block pose dominates latent distance: shifting the block by one pixel displaces the embedding roughly $3\times$ further than an equivalent agent shift, and swapping the block pose alone accounts for $99\%$ of the squared distance between scene pairs.
This clarifies the recent finding of Hu~\etal~\cite{hu2026geometry} that physical state ``barely influences'' LeWM's latent distance (Spearman $0.13$).
We show that this low global correlation stems from a \emph{ceiling effect}, where latent distance correlates strongly with state among nearby scenes (Spearman $0.41$ on the closest $5\%$ of pairs) but saturates across distant pairs.
Moreover, because block pose is encoded along a \emph{curved, distributed manifold} whose local tangent rotates with object position, no single linear weight edit can strip or redirect this information, yet an image-specific perturbation easily can.

We leverage this structure to construct a test-time state-spoofing attack (\cref{sec:method,sec:experiments}; \cref{fig:overview}).
Optimizing a white-box $\ell_\infty$ perturbation of just $\epsilon{=}4/255$ against the public encoder weights on a single image causes an independent state probe to decode the block at an attacker-chosen pose in $73$--$95\%$ of scenes, while the decoded agent position shifts by only a few pixels.
Applying this perturbation offline to the goal image alone drops Push-T planning success from $87\%$ to $32\%$ (\cref{fig:teaser}).
Perturbing the live camera stream is equally damaging, whereas uniform noise at twice the magnitude has no effect.
The attack transfers across all four released LeWM environments (Push-T, Two-Room, Reacher, and OGBench-Cube), where the magnitude of planning failure reflects each task's dependence on the spoofed object, and where on-manifold counterfactual targets prove essential for accurate evaluation (\cref{sec:exp-envs}).

Because the attack leaves model weights untouched and operates purely on input frames, standard model-side checks including SIGReg statistics, clean prediction loss, and nominal task accuracy pass identically to the unperturbed baseline.
While LeWM's surprise monitor can detect isolated perturbed frames, it struggles against temporally consistent spoofing and fails completely against goal-image perturbations, as the forward dynamics model never predicts the goal.
To our knowledge, this is the first demonstration that a Gaussian-regularized JEPA planner can be reliably broken by imperceptible, checkpoint-preserving test-time perturbations.

\noindent\textbf{Contributions.}
\begin{itemize}
\itemsep0.1em
\item We analyze the latent geometry of LeWM on Push-T, showing that block pose dominates latent distance along a curved, high-dimensional manifold and explaining why prior reports of weak state--latent alignment arise from a distance-saturation ceiling effect (\cref{sec:geometry}).
\item We propose a white-box, single-image state-spoofing attack ($\epsilon{=}4/255$) that forces the encoder to report an arbitrary target object pose while preserving non-targeted state variables (\cref{sec:method,sec:exp-spoof}).
\item We evaluate end-to-end MPC planning under the standard LeWM protocol, demonstrating that perturbing either the goal image or live camera observations drops success from $87\%$ to $21$--$35\%$ at $\epsilon{=}4/255$--$8/255$, whereas $2\times$ larger random noise is harmless (\cref{sec:exp-planning}).
\item We demonstrate transfer across all four LeWM benchmarks, showing that planning degradation scales with task sensitivity to the spoofed entity and establishing why on-manifold counterfactuals are required for faithful measurement (\cref{sec:exp-envs}).
\item We show that standard diagnostics like SIGReg, prediction loss, and runtime surprise fail to safeguard the planner, with goal-image attacks bypassing surprise checks by construction (\cref{sec:exp-detection}).
\end{itemize}

%% file: sec/2_related.tex
\section{Related Work}
\label{sec:related}

\paragraph{JEPA world models and planning.}
Visual world models learn environment dynamics to plan over imagined future trajectories~\cite{ha2018worldmodels,hafner2025dreamerv3,hansen2024tdmpc2}.
Joint-Embedding Predictive Architectures (JEPAs) achieve this by predicting target embeddings rather than reconstructing pixels~\cite{lecun2022path,assran2023ijepa,bardes2024vjepa}.
In action-conditioned settings, DINO-WM~\cite{zhou2024dinowm} plans over frozen DINOv2 features~\cite{oquab2024dinov2}, PLDM~\cite{sobal2025pldm} trains end-to-end with a VICReg-inspired objective~\cite{bardes2022vicreg}, and LeWM~\cite{maes2026lewm} trains end-to-end from raw pixels using a single Gaussian regularizer, SIGReg~\cite{balestriero2025lejepa}.
At test time, these frameworks all optimize a latent goal-matching cost using sampling-based solvers such as CEM~\cite{deboer2005cem,pinneri2021icem}.
We center our empirical study on LeWM because its checkpoints and planning code are publicly available, though the core attack surface, control mediated solely through an image encoder and Euclidean latent distance, applies broadly across this family.

\paragraph{Representation geometry for planning.}
Preventing latent collapse does not guarantee a representation geometry conducive to planning, since task-relevant state may be linearly decodable while contributing little to the overall latent distance. 
Hu~\etal~\cite{hu2026geometry} termed this property \emph{metric leverage} and proposed SCALE to calibrate latent distances against task-state distances during training.
Building on their geometric diagnostics, we show that block pose in Push-T actually dominates local latent distances, and we exploit the non-linear structure of this representation to mount a test-time spoofing attack.

\paragraph{Adversarial examples and feature-space attacks.}
Small $\ell_p$-bounded perturbations reliably mislead classifiers~\cite{szegedy2014intriguing,goodfellow2015explaining}, typically via projected gradient descent (PGD)~\cite{carlini2017towards,madry2018towards}.
Instead of targeting class logits, we build on \emph{feature-space} attacks that match internal representations to a guide image~\cite{sabour2016adversarial} or steer generative models at runtime~\cite{bailey2024imagehijacks}.
Here, we attack a world-model encoder by matching the embedding of an on-manifold counterfactual scene to misdirect a downstream planner.
Following standard evaluation practices~\cite{athalye2018obfuscated,carlini2019evaluating,tramer2020adaptive}, we measure spoofing via an external state probe and benchmark against matched-budget noise rather than reporting only the optimized latent loss.

\paragraph{Attacks on control and world models.}
Prior adversarial attacks in control largely target policies that map observations directly to actions, including deep RL agents~\cite{huang2017adversarial,lin2017tactics,zhang2020robust,gleave2020adversarial} and vision-language-action models~\cite{wang2024vla}.
Most relevant to our work, Ria\~no~\etal~\cite{riano2026worldlies} recently introduced a supply-chain backdoor for latent world models (including LeWM), where a compromised checkpoint embeds a trigger-gated action hijack that survives downstream fine-tuning, echoing backdoors in pretrained vision encoders~\cite{gu2019badnets,jia2022badencoder}.
Our threat model is complementary to theirs, as we consider an attacker who cannot modify the checkpoint, training data, or controller, and instead only perturb an input image at test time, leaving the model weights completely clean.
While various input transformations and auxiliary detectors have been explored for classifiers~\cite{cohen2019certified,guo2018countering,metzen2017detecting,xu2018feature}, they remain brittle under adaptive attacks~\cite{carlini2017detected,tramer2020adaptive}.
Because LeWM includes no explicit input defense, we test whether its SIGReg objective and runtime surprise metric provide implicit protection.

%% file: sec/3_background.tex
\section{Background and Threat Model}
\label{sec:background}

\subsection{LeWM and Latent Planning}
\label{sec:bg-lewm}

LeWM pairs a ViT-Tiny encoder $E_\theta$, which maps a $224\times224$ image observation $o_t$ to a latent vector $z_t = E_\theta(o_t)\in\mathbb{R}^{192}$, with an action-conditioned transformer predictor $P_\psi$ that forecasts $\hat z_{t+1} = P_\psi(z_t, a_t)$~\cite{maes2026lewm,dosovitskiy2021vit}.
Both modules are trained jointly on offline trajectories to minimize the one-step prediction loss $\lVert \hat z_{t+1} - z_{t+1}\rVert_2^2$ plus $\lambda\,\mathrm{SIGReg}(Z)$ ($\lambda{=}0.09$).
SIGReg prevents latent collapse by driving the embedding distribution toward an isotropic Gaussian, projecting each batch onto $M{=}1024$ resampled unit directions and penalizing the mean 1D Epps--Pulley normality statistic~\cite{epps1983test,balestriero2025lejepa}.
Note that SIGReg defines no hard pass/fail threshold, operating purely as a batch-averaged training regularizer.

At test time, the frozen model plans via receding-horizon model-predictive control (MPC).
Given the current frame $o_1$ and a goal image $o_g$, LeWM encodes $z_1 = E_\theta(o_1)$ and $z_g = E_\theta(o_g)$ and runs the Cross-Entropy Method (CEM)~\cite{deboer2005cem} to find an action sequence $a_{1:H}$ minimizing the terminal latent cost
\begin{equation}
C(a_{1:H}) = \big\lVert \hat z_{H}(a_{1:H}) - z_g \big\rVert_2^2 ,
\label{eq:cost}
\end{equation}
replanning as new observations arrive~\cite{maes2026lewm}.
Importantly, \cref{eq:cost} routes the entire control objective through the Euclidean geometry of the latent space, anchored by just two encoder passes on $o_1$ and $o_g$.

\subsection{Experimental Setup}
\label{sec:bg-setup}

We evaluate the official \texttt{quentinll/lewm-pusht} checkpoint on Push-T~\cite{maes2026lewm,zhou2024dinowm}, a 2D task where a circular blue agent pushes a grey T-shaped block toward a goal pose.
The arena spans $512\times 512$ units ($224\times 224$ pixels, or $2.28$ units/pixel), with success defined as placing the block within $20$ units and $20^\circ$ of the target.
Following the published protocol (CEM with $300$ samples, $30$ iterations, horizon $5$, frame-skip $5$, $50$-step budget, and goals sampled $25$ steps ahead), we reproduce the reported $96\%$ success rate on the original episode subset.
Because the baseline success rate across independent sets of episodes vary between $85\%$ and $87\%$, every attack is benchmarked against a paired clean run using identical episodes and CEM seeds.
All experiments run the encoder in evaluation mode (\cref{app:impl}), and the other three environments follow the same protocol (\cref{sec:exp-envs}).

\subsection{Threat Model}
\label{sec:bg-threat}

We study a \emph{test-time, input-only} adversary (\cref{tab:threat}, appendix) who perturbs images entering the encoder, the goal image, the live camera frame, or both, within an $\ell_\infty$ bound of $\epsilon/255$ while maintaining valid 8-bit pixel values.
The attacker has white-box access to the public encoder weights $E_\theta$ and can render a counterfactual target scene (an assumption we revisit in \cref{sec:limitations}), but cannot modify the checkpoint, predictor, planner, solver randomness, or task evaluation criteria.
This threat model is orthogonal to the supply-chain backdoor of Ria\~no~\etal~\cite{riano2026worldlies}, which assumes a poisoned checkpoint evaluated on clean inputs.
Here, the checkpoint remains completely unmodified while test-time inputs are perturbed, guaranteeing that any weight-level audit including SIGReg statistics, clean prediction loss, and nominal task accuracy passes by construction.

%% file: sec/4_geometry.tex
\section{Where Task State Lives in LeWM}
\label{sec:geometry}

Before attacking the planner, we characterize how LeWM's latent metric represents task state on Push-T.
This analysis explains why a small pixel perturbation can selectively shift the planner's estimate of the block and why this vulnerability cannot be eliminated by a single linear edit of the encoder.
We probe the encoder using counterfactual renders across $4{,}000$ real scenes, re-rendering each in the simulator with a single state variable altered (agent position, block position, or block angle) and measuring the resulting latent shift relative to the median distance between unrelated scenes ($19.7$).

\subsection{Block Pose Dominates the Latent Metric}
\label{sec:geo-dominates}

\begin{figure}[t]
\centering
\includegraphics[width=\linewidth]{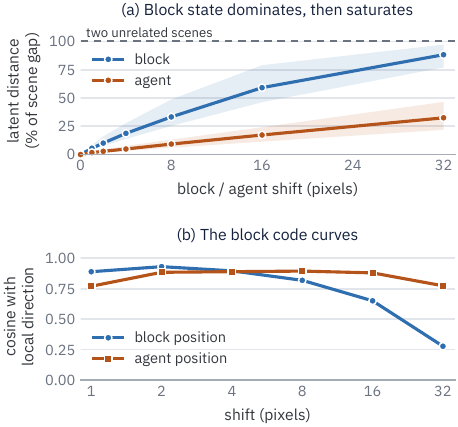}
\caption{\textbf{Latent geometry of task state in LeWM on Push-T.}
(a) Embedding displacement for a single-variable shift as a percentage of the median distance between unrelated scenes ($19.7$), with interquartile bands across scenes.
Block position displaces the embedding far more than agent position and saturates near $16$--$32$ pixels. (b) Cosine similarity between the actual finite displacement and the local tangent predicted from an infinitesimal shift.
Alignment decays rapidly for the block, indicating a curved manifold, whereas the agent trajectory remains nearly linear.}
\label{fig:geometry}
\end{figure}

\paragraph{Sensitivity.} Shifting the block by one pixel displaces the embedding by $0.82$ on average, compared to $0.28$ for a one-pixel agent shift, roughly $3\times$ less (\cref{fig:geometry}a).
Relative to the task's success tolerances, block position is approximately $3$--$4\times$ more influential than agent position.

\paragraph{Swap test.}
Transplanting only the block pose from one scene into another reproduces $99\%$ of the mean squared latent distance between them ($90\%$ from position alone, $61\%$ from angle alone), whereas transplanting the agent position reproduces $44\%$.
Because the planning objective (\cref{eq:cost}) directly minimizes squared Euclidean distance, block pose dominates the cost landscape optimized by the planner, consistent with recent analyses of metric calibration~\cite{hu2026geometry}.

\paragraph{A ceiling, not an absence.}
Hu~\etal~\cite{hu2026geometry} report a Spearman rank correlation of $0.13$ between squared latent distance and squared state distance on Push-T, interpreting this low value as evidence that task state ``barely'' influences latent distance.
We reproduce this global figure but find that it stems primarily from \emph{saturation}.
Latent distance grows rapidly with block displacement at short range before flattening, reaching $59\%$ of the unrelated-scene distance at $16$ pixels and $88\%$ at $32$ pixels (\cref{fig:geometry}a).
Because the latent distance between randomly sampled scene pairs rapidly saturate near this ceiling, distant pairs carry minimal ordering signal and depress the global correlation.
Restricting the evaluation to local neighborhoods yields a correlation of $0.41$ on the closest $5\%$ of pairs and $0.57$ on the closest $0.1\%$, versus $0.11$ on the remainder (\cref{tab:ceiling}, appendix).
Task state is thus strongly represented at short range, the exact regime where the planner refines trajectories near the goal.

\subsection{The Representation Is Curved and Distributed}
\label{sec:geo-curved}

Dominating the metric does not imply that block pose occupies a low-dimensional linear subspace.

\paragraph{No fixed low-dimensional subspace.}
Capturing $90\%$ of the block-position response across scenes requires $52$ of the $\sim$$101$ active latent dimensions ($39$ for block angle, $55$ for agent position).
These subspaces also overlap heavily, with the $52$ block dimensions capturing $60\%$ of the agent response, comparable to the $57\%$ they capture of generic inter-scene variation.
Within any single scene, however, the local displacement directions of the block and agent are nearly orthogonal (median $|\cos|$ of $0.12$--$0.21$).
Thus, while the two objects are locally separable, no global ``block axis'' exists.

\paragraph{The block tangent rotates.}
Comparing the actual latent displacement under a finite shift to the local tangent direction estimated from an infinitesimal shift (\cref{fig:geometry}b), cosine similarity for block translation drops from $0.90$--$0.93$ at $2$--$4$ pixels to $0.82$ at $8$, $0.65$ at $16$, and $0.28$ at $32$ pixels, and from $0.93$ at $5^\circ$ to $0.30$ at $45^\circ$ for rotation.
In contrast, the agent stays near $0.88$ out to $16$ pixels.
The block encoding is therefore strongly \emph{curved}, with its local sensitivity direction rotating as the block moves.

\paragraph{Implication for attacks.}
A curved, distributed manifold cannot be stripped or redirected by a single fixed linear edit of the encoder weights, since a projection that cancels block sensitivity in one scene is misaligned in another.
In auxiliary experiments, global linear edits were either undone by brief predictor re-fitting or flagged by spikes in prediction loss, and even optimal linear concept erasure~\cite{belrose2023leace} leaves block pose nonlinearly decodable.
Because the block and agent are locally orthogonal and the block tangent is well-defined at each state, however, an input perturbation \emph{optimized per image} can selectively spoof the block pose while leaving the agent representation intact.

%% file: sec/5_method.tex
\section{Test-Time State-Spoofing Attack}
\label{sec:method}

\subsection{Objective}
\label{sec:method-obj}

The attacker aims to make the encoder register a target object (here, the block) at a chosen pose while leaving the remaining scene state intact, subject to an imperceptible $\ell_\infty$ bound and valid 8-bit quantization.
Given a clean image $o$ with embedding $z_0 = E_\theta(o)$, the attacker renders a \emph{counterfactual} scene $\tilde o$ identical to $o$ except with the block shifted or rotated by a specified offset, yielding the target embedding $z^\star = E_\theta(\tilde o)$.
The attack then searches for a perturbed image $o'$ within the $\epsilon$-ball around $o$ whose embedding matches $z^\star$:
\begin{equation}
o' = \argmin_{\lVert o' - o\rVert_\infty \le \epsilon}\; \big\lVert E_\theta(o') - z^\star\big\rVert_2^2 .
\label{eq:attack}
\end{equation}
Because $z^\star$ is the embedding of a valid on-manifold scene differing from $o$ only in block pose, matching it forces the encoder to represent the block at the counterfeit pose while preserving the agent and background state.
Although the same formulation can spoof the agent, we focus on the block as the primary task-relevant object (\cref{fig:overview}).

\subsection{Optimization}
\label{sec:method-pgd}

We solve \cref{eq:attack} using $\ell_\infty$ projected gradient descent (PGD)~\cite{madry2018towards} against the frozen public encoder weights, running $100$ sign-gradient steps of size $\epsilon/10$ from a uniform random initialization in the $\epsilon$-ball, followed by $20$ discrete unit steps on the integer lattice so that $o'$ is a valid \texttt{uint8} image with no quantization loss (\cref{app:impl}). 
Optimization takes roughly $0.5$\,s per frame on a single GPU ($\sim$$0.06$\,s amortized over a batch, \cref{tab:cost}) and requires no access to the predictor or planner.

\paragraph{Two deployment points.}
Perturbations can be applied wherever observations enter the encoder (\cref{fig:overview}).
In the \emph{goal-image attack}, only the goal image $o_g$ is perturbed once offline before the episode starts, leaving live camera frames untouched so the planner tracks the true state but optimizes toward a corrupted goal pose.
In the \emph{camera-frame attack}, the goal image remains clean while each live observation encoded at replanning steps are perturbed so that the block appears displaced, causing the planner to mislocalize the block relative to the true goal.
The goal-image variant requires a weaker threat assumption (a single offline edit) and proves harder to detect.

\subsection{Measuring Success}
\label{sec:method-eval}

Rather than evaluating the attack by the latent distance loss it directly optimizes, we use an independent state probe~\cite{alain2017probes}: an MLP trained solely on clean renders to predict agent position, block position, and block angle (median test error of $1.2$ pixels on block position, $1.3^\circ$ on angle, and $2.3$ pixels on agent position).
Because the probe is never accessed during attack optimization, it serves as an external readout of the state encoded in $E_\theta(o')$.
We define \emph{spoof success} as placing the decoded block closer to the target counterfactual pose than the true pose, and we report drift in the decoded agent position to measure selectivity.
For closed-loop planning, we use the simulator's ground-truth success criterion (\cref{sec:bg-setup}).
To distinguish selective spoofing from generic corruption~\cite{carlini2019evaluating,tramer2020adaptive}, we compare against matched-budget uniform integer noise and an \emph{untargeted} PGD baseline that maximizes $\lVert E_\theta(o') - z_0\rVert_2^2$.

%% file: sec/6_experiments.tex
\section{Experiments}
\label{sec:experiments}

We evaluate our attack on Push-T under the standard protocol (\cref{sec:bg-setup}), testing state-spoofing selectivity (\cref{sec:exp-spoof}), closed-loop planning degradation (\cref{sec:exp-planning}), transfer across LeWM's other three benchmarks (\cref{sec:exp-envs}), and evasion of built-in runtime checks (\cref{sec:exp-detection}).

\subsection{The Attack Spoofs the Block Reading}
\label{sec:exp-spoof}

\Cref{tab:spoof} summarizes spoofing accuracy across $520$--$600$ scenes per setting.
At $\epsilon{=}4/255$, the independent probe decodes the block closer to a $+17$-pixel counterfactual target than its true pose in $73\%$ of scenes, rising to $90$--$98\%$ at $\epsilon{=}8/255$ for shifts up to $35$ pixels and rotations up to $90^\circ$.
For moderate shifts, the non-targeted agent position drifts by only a few pixels, on par with the probe's $2.3$-pixel clean test error, confirming that the attack exploits the local separability of block and agent representations (\cref{sec:geo-curved}).
Spoofing the agent instead is even easier ($90\%$ at $\epsilon{=}4/255$ for a $9$-pixel shift) while leaving the decoded block pose intact.
Both controls behave as expected: matched-budget uniform noise shifts the embedding by only $0.06$--$0.42$ (versus $7$--$18$ required for a block spoof) and decoded states by under one pixel, whereas untargeted PGD moves the embedding substantially ($6.4$ at $\epsilon{=}1/255$) but corrupts the agent state rather than selectively spoofing the block.

\begin{table}[t]
    \centering
    \caption{\textbf{Spoof success on Push-T} ($520$--$600$ scenes per row).
    Share of scenes where the independent probe places the target object closer to the counterfactual pose than the true pose, and median drift of the non-targeted object (selectivity).}
    \label{tab:spoof}
    \small
    \setlength{\tabcolsep}{3.5pt}
    \begin{tabular}{@{}llcc@{}}
        \toprule
        Counterfactual scene & $\epsilon$ & spoof succ. & non-target drift \\
        \midrule
        Block $+9$\,px   & $2$/$4$     & $64\%$ / $95\%$          & agent $3$\,px \\
        \rowcolor{gray!12}
        Block $+17$\,px  & $4$/$8$     & $73\%$ / $98\%$          & agent $6$\,px \\
        Block $+35$\,px  & $4$/$8$     & $24\%$ / $90\%$          & agent $16$/$12$\,px \\
        \rowcolor{gray!12}
        Block rot.\ $45^\circ$ & $4$/$8$ & $81\%$ / $95\%$         & agent $9$/$7$\,px \\
        Agent $+9$\,px   & $1$/$2$/$4$ & $54\%$ / $79\%$ / $90\%$ & block $<1$\,px \\
        \midrule
        Uniform noise    & $8$         & $<1\%$                   & all $<1$\,px \\
        \bottomrule
    \end{tabular}
\end{table}

\subsection{Planning Collapses Under the Attack}
\label{sec:exp-planning}

We next inject the perturbation into the MPC loop using the $+17$-pixel block target across $300$ episodes ($6$ seeds $\times\,50$), paired with clean runs on the exact same episodes (\cref{tab:planning}; \cref{fig:planning}).

\paragraph{Goal-image attack.}
Perturbing only the goal image reduces planning success monotonically from $87.0\%$ to $32.3\%$ at $\epsilon{=}4/255$ ($-54.7$ points, paired $95\%$ CI $[-60.7,-48.7]$) and $21.3\%$ at $\epsilon{=}8/255$. 
Feeding the planner a clean render of the counterfactual goal (``perfect'') yields $18.3\%$, showing that $\epsilon{=}4/255$ already achieves most of the possible degradation.
The failure is directed rather than random: across evaluation trials, the fraction of episodes in which the block terminates closer to the spoofed target than the ground-truth goal increase from $6\%$ at $\epsilon{=}1/255$ to $65\%$ at $\epsilon{=}8/255$, while the agent still reaches its target region.
Other counterfactual targets behave similarly (\cref{tab:fakes}), with a $35$-pixel shift dropping success to $7.7\%$ at $\epsilon{=}8/255$ (matching the perfect-fake floor) and a $45^\circ$ rotation dropping it to $25.3\%$.

\begin{table}[t]
    \centering
    \caption{\textbf{Planning under attack on Push-T} (success \%, $300$ episodes, $+17$\,px block target). 
    ``Perfect'' feeds the planner a clean render of the counterfactual target; matched-budget noise is harmless.}
    \label{tab:planning}
    \small
    \setlength{\tabcolsep}{2pt}
    \begin{tabular}{@{}lcccccc@{}}
        \toprule
        Attacked input & clean & $\epsilon{=}1$ & $2$ & $4$ & $8$ & perfect \\
        \midrule
        Goal image   & $87.0$ & $79.0$ & $62.0$ & $32.3$ & $21.3$ & $18.3$ \\
        \rowcolor{gray!12}
        Camera frame & $87.3$ & --     & $49.3$ & $35.3$ & $25.3$ & $27.0$ \\
        \midrule
        Noise $\epsilon{=}8$ & \multicolumn{6}{c}{goal $87.7$\;\; camera $88.3$} \\
        \bottomrule
    \end{tabular}
\end{table}

\begin{table}[t]
    \centering
    \caption{\textbf{Other counterfactual targets on Push-T} (goal-image attack, success \%, $300$ episodes). 
    ``Nearer'' is the share of episodes at $\epsilon{=}8/255$ where the block ends closer to the spoofed pose than the true goal.}
    \label{tab:fakes}
    \small
    \setlength{\tabcolsep}{2.4pt}
    \begin{tabular}{@{}lcccccc@{}}
        \toprule
        Counterfactual scene & $\epsilon{=}1$ & $2$ & $4$ & $8$ & perfect & nearer \\
        \midrule
        Block $+17$\,px & $79.0$ & $62.0$ & $32.3$ & $21.3$ & $18.3$ & $65\%$ \\
        \rowcolor{gray!12}
        Block $+35$\,px & $76.3$ & $57.7$ & $25.0$ & $7.7$ & $7.7$ & $64\%$ \\
        Block rot.\ $45^\circ$ & $82.7$ & $67.3$ & $32.0$ & $25.3$ & $28.0$ & $47\%$ \\
        \bottomrule
    \end{tabular}
\end{table}

\paragraph{Camera-frame attack and controls.}
Perturbing live camera observations during replanning is equally damaging ($35.3\%$ at $\epsilon{=}4/255$, $25.3\%$ at $\epsilon{=}8/255$; idealized floor $27.0\%$).
Uniform noise at $\epsilon{=}8/255$ leaves success unchanged at both injection points, and a clean re-render of the \emph{true} goal matches baseline performance ($85.3\%$), confirming that degradation arises from the optimized perturbation rather than noise or rendering artifacts.

\subsection{The Attack Transfers Across Environments}
\label{sec:exp-envs}

\begin{figure}[t]
    \centering
    \includegraphics[width=\linewidth]{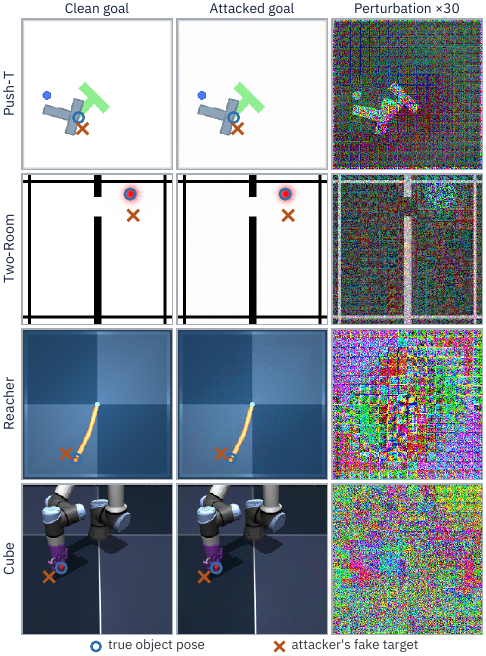}
    \caption{\textbf{Clean versus attacked goal frames across the four LeWM environments} ($\epsilon{=}4/255$; selected by a fixed rule, \cref{app:examples}).
    Residuals are amplified $30\times$. Circles mark the true object pose (block, agent, fingertip, cube) and crosses mark the spoofed target pose; the attack closes $86\%$, $92\%$, $82\%$, and $95\%$ of the squared distance to the target embedding (top to bottom).}
    \label{fig:examples}
\end{figure}

We evaluate the goal-image attack on LeWM's three remaining benchmarks, Two-Room, Reacher, and OGBench-Cube, under the same protocol ($5$ seeds $\times\,50$ episodes), shifting the target entity by roughly twice its goal tolerance ($0.4$\,rad on Reacher, as twice its tolerance is a barely visible $5.7^\circ$; \cref{app:envs}). 
\Cref{fig:examples} shows that the perturbations remain visually imperceptible while driving the embedding nearly all the way to the counterfactual target.
At $\epsilon{=}8/255$, the attack closes $82$--$97\%$ of the squared clean-to-counterfactual latent distance ($90$--$99\%$ at $\epsilon{=}16/255$; \cref{tab:gap}, appendix), and by $\epsilon{=}16/255$ planning success falls to within $3$ points of (or below) the idealized floor across all tasks while uniform noise remains harmless (\cref{tab:envs}).

\begin{table}[t]
    \centering
    \caption{\textbf{Goal-image attack across all four LeWM environments} (success \%; Push-T: $6$ seeds, others $5$, $\times50$ episodes).
    ``Perfect'' uses a dataset-faithful counterfactual render; ``Noise'' is $\epsilon{=}8/255$ uniform noise.}
    \label{tab:envs}
    \small
    \setlength{\tabcolsep}{3.4pt}
    \begin{tabular}{@{}lcccccc@{}}
        \toprule
        Environment & clean & $\epsilon{=}4$ & $8$ & $16$ & perfect & noise \\
        \midrule
        Push-T   & $87.0$ & $32.3$ & $21.3$ & $19.0$ & $18.3$ & $87.7$ \\
        \rowcolor{gray!12}
        Two-Room & $86.0$ & $38.4$ & $30.0$ & $32.4$ & $38.4$ & $87.2$ \\
        Reacher  & $61.2$ & $29.6$ & $26.0$ & $20.4$ & $17.6$ & $60.4$ \\
        \rowcolor{gray!12}
        Cube     & $70.0$ & $53.6$ & $53.2$ & $50.0$ & $55.6$ & $70.8$ \\
        \bottomrule
    \end{tabular}
\end{table}

\begin{table}[t]
    \centering
    \caption{\textbf{Task sensitivity.}
    Share of episodes already satisfying the goal at initialization, and goal-image success (\%) on the remaining episodes ($n{=}299/229/249/67$).
    Last row: the $183$ Cube episodes already met at start.}
    \label{tab:sens}
    \small
    \setlength{\tabcolsep}{3.6pt}
    \begin{tabular}{@{}lcccc@{}}
        \toprule
        Environment & met at start & clean & $\epsilon{=}8$ & perfect \\
        \midrule
        Push-T   & $0.3\%$  & $87.0$ & $21.1$ & $18.1$ \\
        \rowcolor{gray!12}
        Two-Room & $8.4\%$  & $84.7$ & $23.6$ & $32.8$ \\
        Reacher  & $0.4\%$  & $61.0$ & $25.7$ & $17.3$ \\
        \rowcolor{gray!12}
        Cube     & $73.2\%$ & $38.8$ & $28.4$ & $26.9$ \\
        \quad already met & -- & $81.4$ & $62.3$ & $66.1$ \\
        \bottomrule
    \end{tabular}
\end{table}

\paragraph{Planning collapse tracks task sensitivity.}
In tasks where the planner directly controls the spoofed variable, agent position in Two-Room and arm configuration in Reacher, planning drops sharply at $\epsilon{=}8/255$ ($86.0\%\to30.0\%$ and $61.2\%\to26.0\%$), matching Push-T.
On Cube, the encoder is spoofed equally well ($97\%$ distance closed), yet overall success drops only from $70.0\%$ to $53.2\%$ (paired $95\%$ CI $[-23,-11]$ points vs.\ $[-71,-60]$ on Push-T).
\Cref{tab:sens} explains why: in $73.2\%$ of Cube episodes, the goal cube already lies within the $4$\,cm tolerance at the start (vs.\ $<9\%$ elsewhere), and on the remaining episodes the clean planner succeeds only $38.8\%$ of the time, leaving little margin for a fake goal to degrade (an unperturbed fake goal yields $26.9\%$).
Even when the cube starts inside the goal tolerance, a spoofed goal still costs $15$--$19$ points ($81.4\%\to62.3\%$).
Latent spoofing is thus \emph{necessary but not sufficient} for catastrophic failure, as downstream impact is bounded by the task's reliance on the spoofed variable.

\paragraph{On-manifold counterfactuals are required.}
For the MuJoCo benchmarks, naively re-rendering a state can push the target image off the training manifold.
Our standalone Reacher renderer draws the DM-Control checker floor in a shifted phase relative to the dataset, causing the encoder to map an \emph{identical} state $223$ squared units away from the dataset frame, more than a real $0.4$\,rad arm move ($193$).
Using this off-manifold render fabricates an artificial collapse (replacing a goal frame with its zero-shift re-render drops clean success from $56\%$ to $27$--$30\%$) and misleads the attack (\cref{app:envs}). 
We therefore draw MuJoCo counterfactuals on-manifold: a real same-episode frame with the arm displaced for Reacher, and a re-rendered simulator state with only the cube moved for Cube (whose renderer matches the dataset).
An independent probe trained on real frames then decodes the attacked Reacher arm $20^\circ$ from its true angle at $\epsilon{=}8/255$ (tolerance $2.9^\circ$).

\subsection{Built-In Diagnostics Do Not Catch the Attack}
\label{sec:exp-detection}

Because the checkpoint is unmodified, SIGReg, clean prediction loss, and nominal task success match the official release.
At runtime, LeWM's only built-in monitor is \emph{surprise}, the discrepancy between the predicted next latent $\hat z_{t+1}$ and the encoded observation $z_{t+1}$ (\cref{tab:detection}).
While surprise flags a \emph{single} spoofed frame ($100\%$ of clips), it catches only $42$--$59\%$ of clips at a $5\%$ false-positive rate when frames are perturbed consistently across the rollout (AUROC $0.76$--$0.91$).
Crucially, the \emph{goal-image attack escapes this check entirely} because the forward model never predicts the goal image.
Evaluating SIGReg on test batches is similarly ineffective, flagging $49$--$100\%$ of clean, low-motion trajectories (\cref{app:detection}).

\begin{table}[t]
    \centering
    \caption{\textbf{Surprise as a runtime detector} (Push-T, $498$ clips, $+17$\,px block target; TPR reported at $5\%$ FPR). An isolated spoofed frame is reliably flagged, a temporally consistent camera attack is missed roughly half the time, and goal-image attacks trigger no surprise by construction.}
    \label{tab:detection}
    \small
    \setlength{\tabcolsep}{3.5pt}
    \begin{tabular}{@{}lccc@{}}
        \toprule
        Attacked input ($\epsilon{=}2/4/8$) & Surprise ($\times$) & AUROC & TPR@5\% \\
        \midrule
        Single frame & $23$/$70$/$116\times$ & $1.00$ & $100\%$ \\
        \rowcolor{gray!12}
        All frames   & $3.4$/$5.4$/$5.6\times$ & $.84$/$.89$/$.91$ & $44$/$57$/$59\%$ \\
        Goal image (any $\epsilon$) & \multicolumn{3}{c}{not predicted (no surprise)} \\
        \bottomrule
    \end{tabular}
\end{table}

%% file: sec/7_conclusion.tex
\section{Discussion and Limitations}
\label{sec:limitations}

\paragraph{Assumptions.}
Our attack operates white-box against the public encoder weights and perturbs digital images entering the encoder rather than executing a physical-world attack.
Generating the counterfactual target also relies on rendering a modified simulator state, which an external attacker would have to estimate from observations.
Given that an MLP probe accurately decodes physical state from clean embeddings (\cref{sec:method-eval}), we expect such estimation to be viable, though fully black-box and transfer attacks remain open for future work.

\paragraph{Scope.}
While we evaluate all four LeWM benchmarks, our cross-environment study focuses on the goal-image attack (with camera-frame experiments on Push-T and Two-Room in \cref{app:envs}), tests a single target offset per environment (Reacher at $0.4$\,rad rather than twice its tolerance), and restrict the detailed geometry analysis to Push-T.
The variation across benchmarks also nuances our central finding (\cref{sec:exp-envs}), demonstrating that a complete latent spoof only collapses planning when the task actively depends on the spoofed variable. 
Lastly, we evaluate LeWM's existing diagnostics as detectors but do not propose a new defense.

\paragraph{Defenses and the regularizer.}
Candidate input-level defenses include input smoothing or quantization~\cite{xu2018feature,guo2018countering,cohen2019certified}, checking latent consistency between the goal image and initial camera frame to catch goal-image spoofing, and adversarial training of the encoder.
Because auxiliary detectors are routinely broken by adaptive attacks~\cite{carlini2017detected,tramer2020adaptive}, any defense must be benchmarked against a defense-aware adversary, and whether geometry-calibrated encoders such as SCALE~\cite{hu2026geometry} are more or less vulnerable remains an open question.
Fundamentally, SIGReg regularizes the global batch distribution of embeddings rather than the encoder's local sensitivity to individual inputs, confirming that anti-collapse regularization provides no robustness at the pixel-to-latent interface.
This complements the supply-chain backdoor of Ria\~no~\etal~\cite{riano2026worldlies}, showing that LeWM can be compromised either by a poisoned checkpoint on clean inputs or by a clean checkpoint on a single perturbed input.

\section{Conclusion}
\label{sec:conclusion}

We demonstrated that a Gaussian-regularized JEPA planner can be driven from $87\%$ to $32\%$ success on Push-T by an imperceptible $\epsilon{=}4/255$ perturbation of a single input image, while leaving the official checkpoint, and every model-side health check, completely untouched.
The attack transfers across LeWM's other three benchmarks, where planning degradation scales with task dependence on the spoofed object.
This vulnerability stems directly from the model's latent geometry: block pose dominates the planning metric but is encoded along a curved, distributed manifold that prevents global linear erasure while permitting local per-image redirection.
Our findings highlight the image-to-latent interface as a critical vulnerability in reusable JEPA world models and motivate input-level defenses that are robust by design.

%% file: sec/8_appendix.tex
\section{Threat Model Comparison}
\label{app:threat}

\Cref{tab:threat} contrasts our test-time, input-only threat model (\cref{sec:bg-threat}) against the supply-chain checkpoint backdoor of Ria\~no~\etal~\cite{riano2026worldlies}, summarizing how the assumptions on attacker capability differs between the two settings.

\begin{table}[h]
    \centering
    \caption{\textbf{Comparison of threat models.} We contrast our test-time, input-only adversary against the supply-chain checkpoint backdoor of Ria\~no~\etal~\cite{riano2026worldlies}.
    A ``\cmark'' indicates components modified or controlled by the adversary.}
    \label{tab:threat}
    \small
    \setlength{\tabcolsep}{4pt}
    \begin{tabular}{@{}lcc@{}}
        \toprule
        Attacker controls & Backdoor~\cite{riano2026worldlies} & \textbf{Ours} \\
        \midrule
        Released checkpoint (weights)     & \cmark & \xmark \\
        Training data / pipeline          & \xmark & \xmark \\
        Downstream controller / planner   & \xmark & \xmark \\
        Single input image at test time   & \xmark & \cmark \\
        \midrule
        Passes clean checkpoint audit     & partial & \textbf{by constr.} \\
        \bottomrule
    \end{tabular}
\end{table}

\section{Implementation Details}
\label{app:impl}

\paragraph{Evaluation protocol.}
All planning experiments execute the vision encoder in standard evaluation mode.
Because LeWM's projector head incorporates BatchNorm layers, its latent embeddings differ between training and evaluation modes; we use evaluation mode throughout to match the deployed inference configuration. 
Evaluation episodes are sampled identically to the official LeWM benchmark code (using the same dataset starting indices with goal observations sampled $25$ steps ahead).
Every attack condition replays the exact same episodes using identical CEM random seeds, ensuring strictly paired comparisons.
Push-T evaluations use seeds $42$--$47$ ($6\times50=300$ episodes), while the remaining three benchmarks use seeds $42$--$46$ ($5\times50=250$ episodes).

\paragraph{Attack optimization.}
We optimize our objective using $100$ $\ell_\infty$ PGD sign-gradient steps of size $\epsilon/10$ from a uniform random initialization within the $\epsilon$-ball, projecting back onto both the $\epsilon$-ball and the valid $[0,255]$ pixel range after each iteration.
The resulting continuous iterate is then rounded to the nearest integer and refined with $20$ discrete unit-sign steps directly on the integer lattice, retaining the candidate perturbation across all iterations that achieve the lowest latent matching loss so that the final output is a valid 8-bit image.
\Cref{tab:cost} ablates the optimization budget without retuning the step size.
Under our default $100+20$ schedule, attacking a single frame takes $0.46$\,s on one GPU ($121$ forward-backward encoder passes at $\sim$$4$\,ms per pass), $3.39$\,s across $8$ CPU threads, and $57$\,ms per frame when amortized across a batch of $250$ images.
Reducing the schedule to $20+5$ steps ($26$ passes, $0.10$\,s per image) preserves nearly all of the spoofing effectiveness.

\begin{table}[h]
    \centering
    \caption{\textbf{Computational cost versus PGD iteration budget} (Push-T, $+17$\,px block target, $\epsilon{=}4/255$, $599$ scenes).
    ``Steps'' denotes continuous sign-gradient steps in the $\epsilon$-ball plus discrete unit steps on the integer lattice (top row corresponds to our default configuration).
    Runtime is reported for a single image on one GPU (NVIDIA RTX PRO 5000), across $8$ CPU threads, and amortized per image over a batch of $250$ on the GPU.
    ``Spoof succ.'' matches the definition in \cref{tab:spoof}.}
    \label{tab:cost}
    \small
    \setlength{\tabcolsep}{3.4pt}
    \begin{tabular}{@{}lccccc@{}}
        \toprule
        Steps & passes & GPU (s) & batch (ms) & CPU (s) & spoof succ. \\
        \midrule
        $100+20$ & $121$ & $0.46$ & $57$ & $3.39$ & $73\%$ \\
        \rowcolor{gray!12}
        $50+20$  & $71$  & $0.27$ & $35$ & $1.83$ & $73\%$ \\
        $20+20$  & $41$  & $0.19$ & $20$ & $1.21$ & $71\%$ \\
        \rowcolor{gray!12}
        $10+20$  & $31$  & $0.12$ & $15$ & $0.91$ & $70\%$ \\
        $20+5$   & $26$  & $0.10$ & $13$ & $0.76$ & $70\%$ \\
        \rowcolor{gray!12}
        $10+5$   & $16$  & $0.07$ & $8$  & $0.44$ & $62\%$ \\
        $5+5$    & $11$  & $0.04$ & $5$  & $0.29$ & $57\%$ \\
        \bottomrule
    \end{tabular}
\end{table}

\paragraph{State probe architecture (Push-T).}
Our external state probe is a two-hidden-layer MLP ($256$ hidden units per layer with GELU activations) that maps the $192$-dimensional latent embedding to agent position, block position, and block orientation (parameterized as $(\sin\theta, \cos\theta)$).
We train the probe on approximately $20{,}000$ clean simulator renders ($90/10$ train/validation split) using AdamW with learning rate $10^{-3}$, weight decay $10^{-4}$, and a cosine annealing schedule over $3{,}000$ steps at a batch size of $2{,}048$.
On held-out clean data, the probe achieves median errors of $1.2$ pixels on block position, $1.3^\circ$ on block angle, and $2.3$ pixels on agent position.
The probe weights are kept strictly isolated from the attack optimization loop.
The Reacher joint-angle probe in \cref{app:envs} uses the exact same architecture and training hyperparameters on $9{,}000$ real dataset frames.

\section{Geometry Details}
\label{app:geometry}

\Cref{tab:ceiling} provides the quantitative breakdown showing that the weak global correlation between latent distance and state distance discussed in \cref{sec:geo-dominates} is driven by distance saturation. 
When isolating block translation by restricting the evaluation to scene pairs with similar block orientations, the rank correlation between latent and physical distances remain strong up to roughly $16$ pixels of separation (Spearman $\rho_S = 0.47$) before collapsing to near zero beyond that threshold ($\rho_S = 0.02$).

\begin{table}[h]
    \centering
    \caption{\textbf{Distance saturation in LeWM's latent metric.}
    We report the Spearman rank correlation ($\rho_S$) between squared latent distance and squared ground-truth state distance across $6.3$M clean Push-T scene pairs, stratified by physical state separation (alongside median block separation and median Euclidean latent distance).
    Results are consistent across random seeds.}
    \label{tab:ceiling}
    \small
    \setlength{\tabcolsep}{5pt}
    \begin{tabular}{@{}lccc@{}}
        \toprule
        Scene pair subset & block sep.\ & latent dist.\ & Spearman $\rho_S$ \\
        \midrule
        All pairs        & 51\,px & 19.7 & 0.16 \\
        Closest $0.1\%$  & \phantom{0}4\,px  & \phantom{0}8.0 & \textbf{0.57} \\
        Closest $5\%$    & 14\,px & 18.5 & \textbf{0.41} \\
        Farthest $95\%$  & 53\,px & 19.7 & 0.11 \\
        \bottomrule
    \end{tabular}
\end{table}

\section{Planning Success versus Budget}
\label{app:planning}

\begin{figure}[h]
    \centering
    \includegraphics[width=\linewidth]{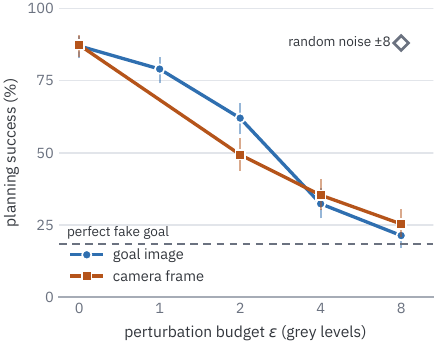}
    \caption{\textbf{Closed-loop planning success versus perturbation budget $\epsilon$ on Push-T} ($300$ episodes; error bars denote $95\%$ Wilson confidence intervals).
    Both the goal-image and camera-frame attacks degrade performance monotonically toward the unconstrained counterfactual floor, while uniform random noise of equal magnitude has no effect.}
    \label{fig:planning}
\end{figure}

\Cref{fig:planning} visualizes the dose-response curves corresponding to \cref{tab:planning}.
Perturbing either the static goal image or the incoming camera frames at replanning steps steadily drive Push-T task completion down toward the idealized ``perfect counterfactual goal'' floor as $\epsilon$ increases, whereas matched-budget uniform random noise leaves planning performance untouched.

\section{Detection by Surprise}
\label{app:detection}

LeWM's primary runtime monitoring signal is \emph{surprise}, the squared Euclidean error between the forward predictor's one-step forecast $\hat z_{t+1}$ and the actually observed embedding $z_{t+1}$, which the original authors propose as a violation-of-expectation detector.
We evaluate surprise across $498$ trajectory clips under a counterfactual attack shifting the block by $17$ pixels (\cref{tab:detection}, main text).
Surprise reliably flags an isolated spoofed frame injected into an otherwise clean rollout, as a single-step shift mimics an instantaneous object teleportation and elevates prediction error by $23$--$132\times$ across $100\%$ of clips.
However, when an adversary perturbs every incoming frame toward a temporally consistent counterfactual trajectory, the resulting latent sequence remains dynamically coherent: surprise increases by only $2.8$--$5.6\times$ and flags just $42$--$59\%$ of clips at a $5\%$ false-positive rate (AUROC $0.76$--$0.91$). 
More fundamentally, the \emph{goal-image attack bypasses this mechanism by construction}, because the forward dynamics model never predicts the static goal embedding and thus computes no prediction residual for it.

We also evaluated whether test-time SIGReg statistics can serve as an input-level anomaly monitor, but found that the normality score across batches of perturbed observations provide no discriminative signal beyond the natural variance of the counterfactual scenes themselves.
In practice, SIGReg flags $49$--$100\%$ of clean trajectory batches that exhibit low motion diversity, rendering it unable to distinguish adversarial perturbations from benign distribution shifts.
Consequently, the goal-image attack evades all built-in diagnostics offered by the model.

\section{Cross-Environment Details}
\label{app:envs}

\paragraph{Evaluation protocol.}
\Cref{tab:protocol} summarizes the manipulated target entity, nominal goal tolerance, counterfactual shift magnitude, and counterfactual generation method for each benchmark.
We set the counterfactual displacement to twice the task's success tolerance across all environments except Reacher, where twice the nominal tolerance ($0.1$\,rad, or $5.7^\circ$) produces a barely perceptible arm movement; we therefore use a $0.4$\,rad ($23^\circ$) shift on Reacher.
Closed-loop planning uses each benchmark's official pretrained LeWM checkpoint and hyperparameter configuration.
Across all tables, ``perfect'' feeds the unperturbed counterfactual target image directly to the planner, while ``noise'' applies uniform integer perturbations in $[-\epsilon, +\epsilon]$ at $\epsilon{=}8/255$ to the goal image.

\begin{table}[h]
    \centering
    \caption{\textbf{Cross-environment attack protocol.}
    Counterfactual target sources: \emph{render} uses our standalone simulator renderer; \emph{real frame} selects an on-manifold dataset frame from the same episode with a displaced arm configuration; \emph{goal state} re-renders the goal frame's exact simulator state after translating only the cube.}
    \label{tab:protocol}
    \small
    \setlength{\tabcolsep}{2.6pt}
    \begin{tabular}{@{}lllll@{}}
        \toprule
        Environment & entity & tolerance & target shift & target source \\
        \midrule
        Push-T   & block & $8.8$\,px   & $17.5$\,px & render \\
        Two-Room & agent & $16$\,px    & $32$\,px   & render \\
        Reacher  & arm   & $0.05$\,rad & $0.4$\,rad & real frame \\
        Cube     & cube  & $4$\,cm     & $8$\,cm    & goal state \\
        \bottomrule
    \end{tabular}
\end{table}

\paragraph{Renderer fidelity.}
To verify whether standalone simulator renders lie on the training manifold, we measure the squared latent distance between real dataset frames and our simulator's re-render of the \emph{exact same} physical state, comparing this gap against the latent distance to the displaced counterfactual target (\cref{tab:fidelity}; $50$ goal frames per benchmark, seed $42$).
For Push-T, Two-Room, and Cube, simulator re-renders match the dataset manifold closely, with a baseline rendering gap of $\le 0.34$ squared units that is negligible compared to the $>200$-unit distance induced by the counterfactual state shift.
On Reacher, however, the standalone DeepMind Control Suite renderer draws the checkerboard floor with a shifted texture phase relative to the released dataset, causing the encoder to map an identical arm configuration $223$ squared units away from the dataset frame, exceeding the $193$-unit shift produced by a true $0.4$\,rad arm displacement.

\begin{table}[h]
    \centering
    \caption{\textbf{Simulator rendering fidelity.}
    Median squared Euclidean distance in latent space between a dataset frame and our standalone re-render of the identical state, compared against the median squared distance between the clean goal frame and the displaced counterfactual target ($50$ goal frames per benchmark).}
    \label{tab:fidelity}
    \small
    \setlength{\tabcolsep}{4pt}
    \begin{tabular}{@{}lcc@{}}
        \toprule
        Environment & re-render vs.\ dataset & clean goal $\to$ target \\
        \midrule
        Push-T   & $0.34$         & $278$ \\
        Two-Room & $<10^{-4}$     & $390$ \\
        Cube     & $0.005$        & $222$ \\
        Reacher  & $\mathbf{223}$ & $193$ \\
        \bottomrule
    \end{tabular}
\end{table}

\paragraph{Why Reacher requires on-manifold dataset targets.}
Using an off-manifold re-render as the counterfactual target distorts evaluation in two ways.
First, it fabricates artificial planning failures: simply replacing the dataset goal image with our standalone re-render at \emph{zero} arm displacement drops Reacher planning success from $56.0\%$ to $27.3\%$ (under the goal-image protocol) and $30.0\%$ (under the camera-frame protocol) across $3$ seeds $\times\,50$ episodes, meaning that the vast majority of the apparent performance collapse stems from the rendering discrepancy rather than arm mislocalization.
Second, evaluating the attack exclusively on simulator re-renders, where both the clean observation and the counterfactual target is generated by the standalone renderer, understates attack effectiveness: at $\epsilon{=}8/255$, PGD closes only $60\%$ of the squared latent gap on re-rendered pairs (and $56\%$ in a preliminary $400$-scene probe study where the probe decoded the arm closer to the target in $55\%$ of trials), compared to closing $82\%$ of the gap when attacking real dataset frames ($81\%$ when targeting a re-rendered goal from a real frame).
We therefore source all Reacher counterfactual targets from real dataset frames throughout our experiments.

\paragraph{Joint-angle probe on real Reacher frames.}
To verify that our attack genuinely shifts the encoded arm state on Reacher, we train an MLP joint-angle probe on $9{,}000$ clean dataset frames (achieving a median held-out error of $0.26^\circ$) and attack $80$ real goal frames toward displaced dataset frames ($22.9^\circ$ median displacement, vs.\ a $2.9^\circ$ task tolerance).
As shown in \cref{tab:reader}, the attack shifts the decoded joint angles by $20.1^\circ$ away from the true configuration at $\epsilon{=}8/255$, placing the arm closer to the counterfactual target in $72.5\%$ of scenes, and brings the decoded pose to within $4.9^\circ$ of the target configuration at $\epsilon{=}16/255$.

\begin{table}[h]
    \centering
    \caption{\textbf{Reacher joint-angle probe evaluation on real dataset frames} ($80$ goal frames across $\ell_\infty$ budgets $\epsilon$).
    Angular errors denote the maximum per-joint deviation in degrees (baseline clean probe error is $0.26^\circ$).}
    \label{tab:reader}
    \small
    \setlength{\tabcolsep}{4pt}
    \begin{tabular}{@{}lcccc@{}}
        \toprule
        $\epsilon$ & nearer target & err.\ to target & err.\ to true & gap closed \\
        \midrule
        $2$  & $56.3\%$ & $15.6^\circ$ & $17.6^\circ$ & $0.75$ \\
        $4$  & $58.8\%$ & $13.4^\circ$ & $16.0^\circ$ & $0.82$ \\
        $8$  & $72.5\%$ & $11.9^\circ$ & $20.1^\circ$ & $0.80$ \\
        $16$ & $98.8\%$ & $4.9^\circ$  & $20.8^\circ$ & $0.95$ \\
        \bottomrule
    \end{tabular}
\end{table}

\paragraph{Task sensitivity analysis on Cube.}
\Cref{tab:sens} evaluates each benchmark's ground-truth goal criterion directly on the initial start and goal states of the evaluation episodes (Push-T: agent and block position within $20$ units and angle within $\pi/9$; Two-Room: agent within $16$\,px; Reacher: both joints within $0.05$\,rad; Cube: cube position within $4$\,cm) and partitions per-episode planning outcomes according to whether the goal tolerance is already satisfied at $t{=}0$ (which occurs in only a single episode each on Push-T and Reacher).
On Cube, the $\epsilon{=}8/255$ attack reduces success by $-10.4$ percentage points across the $67$ active episodes that require moving the cube (paired $95\%$ CI $[-25.4,+4.5]$) and by $-19.1$ points across the $183$ episodes initialized inside the goal region ($[-25.7,-12.6]$); an unperturbed counterfactual goal reduces success by $-11.9$ and $-15.3$ points on these respective splits.
Extending the episode horizon does not amplify the degradation: when sampling goals $75$ steps ahead ($2$ seeds $\times\,30$ episodes), Cube planning success moves from $71.7\%$ (clean) to $65.0\%$ under both the unconstrained counterfactual goal and the $\epsilon{=}8/255$ attack.

\paragraph{Latent distance gap closed across budgets.}
\Cref{tab:gap} reports the fraction of squared clean-to-counterfactual latent distance closed by the goal-image attack across budgets, corresponding to the summary ranges cited in the main text ($0.82$--$0.97$ at $\epsilon{=}8/255$ and $0.90$--$0.99$ at $\epsilon{=}16/255$).
Using the identical metric across all four benchmarks, the fraction of distance closed increases monotonically with $\epsilon$ everywhere except Two-Room, where it plateaus between $0.86$ and $0.91$.

\begin{table}[h]
    \centering
    \caption{\textbf{Fraction of squared latent distance closed by the goal-image attack}, defined as $1-\lVert z_{\mathrm{adv}}-z^\star\rVert_2^2/\lVert z_{0}-z^\star\rVert_2^2$ (median across episodes per seed, averaged across seeds; $6$ seeds on Push-T, $5$ on others).}
    \label{tab:gap}
    \small
    \setlength{\tabcolsep}{5pt}
    \begin{tabular}{@{}lcccc@{}}
        \toprule
        Environment & $\epsilon{=}2$ & $4$ & $8$ & $16$ \\
        \midrule
        Push-T   & $.52$ & $.86$ & $.97$ & $.99$ \\
        \rowcolor{gray!12}
        Two-Room & $.86$ & $.91$ & $.86$ & $.90$ \\
        Reacher  & $.79$ & $.82$ & $.82$ & $.95$ \\
        \rowcolor{gray!12}
        Cube     & $.93$ & $.96$ & $.97$ & $.99$ \\
        \bottomrule
    \end{tabular}
\end{table}

\paragraph{Additional budgets and Two-Room camera-frame attack.}
At a smaller budget of $\epsilon{=}2/255$, the goal-image attack reduces planning success to $62.0\%$ on Push-T, $43.6\%$ on Two-Room, $30.0\%$ on Reacher, and $57.2\%$ on Cube.
Note that Two-Room success is non-monotonic between $\epsilon{=}8/255$ ($30.0\%$) and $\epsilon{=}16/255$ ($32.4\%$), a minor fluctuation well within seed-to-seed variance.
Applying the live camera-frame attack to Two-Room ($5$ seeds $\times\,50$ episodes) reduces planning success from $86.0\%$ (clean) to $55.2\%$, $46.4\%$, and $48.0\%$ at $\epsilon{=}2/255$, $4/255$, and $8/255$, respectively (compared to $44.8\%$ under an unperturbed counterfactual observation stream and $86.4\%$ under $\epsilon{=}8/255$ uniform noise).

\section{Example Selection Protocol}
\label{app:examples}

\paragraph{Qualitative attack examples (\cref{fig:examples}).}
To avoid cherry-picking, each environment in \cref{fig:examples} is illustrated using a single representative frame chosen via a deterministic rule without visual inspection: among the candidate evaluation frames, we select the sample whose fraction of squared latent distances closed at $\epsilon{=}4/255$ lie closest to the candidate median.
For Two-Room, Reacher, and Cube, the candidate pool comprises the first $16$ planning-evaluation goal frames from seed $42$ (with Cube restricted to the $10$ frames in which the block rests on the tabletop so the counterfactual shift does not leave the cube suspended mid-air).
For Push-T, the candidate pool comprises the $6$ stored scenes from the state-probe evaluation in \cref{sec:exp-spoof} (rendered scenes with a $+17.5$-pixel block shift).
The median fraction of latent distance closed across the candidate pools is $0.87$, $0.91$, $0.80$, and $0.96$ for Push-T, Two-Room, Reacher, and Cube, respectively, and the selected frames close $0.86$, $0.92$, $0.82$, and $0.95$.
Across all four visualizations, the maximum pixel modification is bounded by $\epsilon{=}4/255$.

\paragraph{Teaser episode selection (\cref{fig:teaser}).}
To select the rollout shown in \cref{fig:teaser}, we execute the $50$ Push-T evaluation episodes from seed $42$ under both the clean goal image and the $\epsilon{=}4/255$ perturbed goal image (achieving $94\%$ clean success vs.\ $30\%$ attacked success on this seed, whereas the bar chart in the figure reports the $6$-seed mean).
We filter for the $15$ episodes in which the clean rollout succeeds, the attacked rollout fails, and the terminal block position under attack lands within the task's $20$-unit success threshold of the spoofed target pose.
We then select the median episode ranked by the terminal distance margin (distance to the true goal minus distance to the spoofed target), in which the block finishes $6.8$ pixels from the spoofed pose and $12.8$ pixels from the true goal.